# FROM SILENT SIGNALS TO NATURAL LANGUAGE: A DUAL-STAGE TRANSFORMER-LLM APPROACH

*Nithyashree Sivasubramaniam*

*Independent Researcher, Coimbatore, India*

nithyashree2k@gmail.com

## ABSTRACT

Silent Speech Interfaces (SSIs) have gained attention for their ability to generate intelligible speech from non-acoustic signals. While significant progress has been made in advancing speech generation pipelines, limited work has addressed the recognition and downstream processing of synthesized speech, which often suffers from phonetic ambiguity and noise. To overcome these challenges, we propose an enhanced automatic speech recognition framework that combines a transformer-based acoustic model with a large language model (LLM) for post-processing. The transformer captures full utterance context, while the LLM ensures linguistic consistency. Experimental results show a 16% relative and 6% absolute reduction in word error rate (WER) over a 36% baseline, demonstrating substantial improvements in intelligibility for silent speech interfaces.

*Index Terms*— Silent speech interface, speech recognition, transformer, large language model, phoneme ambiguity.

## 1. INTRODUCTION

Silent Speech Interfaces (SSIs) provide an alternative communication pathway by enabling speech synthesis without vocalization. Instead of relying on acoustic signals, SSIs exploit non-vocal modalities that capture articulatory and physiological activity underlying speech production. A variety of input sources have been investigated, including surface electromyography (EMG), ultrasound tongue imaging (UTI), electromagnetic articulography (EMA), and visual cues from lip movements. These modalities offer complementary perspectives on articulatory dynamics, positioning SSIs as a promising solution for communication in scenarios where acoustic speech is impaired.

Recent advances in SSI have introduced convolutional-transducer models combined with transformer encoders and auxiliary phoneme prediction losses, achieving notable improvements in intelligibility. The input signals typically undergo feature extraction to capture articulatory or physiological attributes that preserve discriminative information necessary for speech generation. These features are commonly transformed into acoustic representations, such as spectrograms or mel-frequency cepstral coefficients (MFCCs), using deep learning architectures that include recurrent, convolutional, and transformer-based networks. The final waveform is reconstructed by neural vocoders that synthesize high-quality speech from these intermediate representations. A critical step in this process is aligning silent EMG signals with their vocalized counterparts, as they are often recorded separately. Dynamic Time Warping (DTW) has been widely used to establish monotonic alignments between predicted acoustic features and reference speech features, enabling effective training despite temporal mismatches. Nevertheless, EMG-based systems remain constrained by the inherent ambiguity of muscle activations, as phonetically distinct sounds can produce similar surface EMG patterns. Baseline systems report word error rates (WER) exceeding 36%, reflecting the difficulty of resolving such ambiguities.

More recent work has explored accent variability through optimized transduction and meta-learning approaches, yet consistent intelligibility across large vocabularies and conversational contexts remains an open challenge. These limitations highlight the need for frameworks that integrate both robust alignment and context-aware recognition. The absence of complete acoustic and articulatory cues in silent speech remains a fundamental limitation that cannot be fully addressed through modality-specific preprocessing, handcrafted feature extraction, or improvements in speech synthesis alone. These challenges motivate the need for approaches that not only generate speech but also mitigate ambiguities at the recognition and language levels.

In parallel, the broader speech community has increasingly employed large language models (LLMs) for generative error correction (GER), where ASR-transcribed outputs are refined to reduce recognition errors. For instance, recent research introduced a generative paradigm that leverages N-best ASR hypothesis lists, achieving significant transcription refinements. Building on this, recent research has demonstrated that task-activating prompting can enable LLMs to perform ASR rescoring and correction effectively without fine-tuning. Other studies have further shown that GPT-4 can enhance transcription accuracy through in-context learning on datasets such as Aishell-1 and LibriSpeech. More recently, LLMs have also been applied to multilingual ASR correction, where consistent WER reductions were observed across diverse languages.

Collectively, these findings underscore the capability of LLMs to enforce linguistic consistency and improve transcription fluency, even under noisy or domain-shifted conditions. For SSIs, where phoneme-level ambiguities frequently propagate into recognition errors, such correction mechanisms are particularly valuable. Unlike acoustic-only improvements, an LLM-based correction module directly addresses the linguistic and semantic layers of intelligibility.

While LLM-based correction has been explored in conventional ASR, to the best of our knowledge, it has not been systematically applied in the SSI domain, where input signals are substantially more ambiguous and traditional acoustic refinements are insufficient. This work is the first to demonstrate that integrating an LLM correction stage into an SSI pipeline can substantially reduce WER and improve intelligibility. To this end, we propose a two-stage framework designed to enhance intelligibility and robustness in SSI-based recognition (Fig. 1). First, we replace recurrent architectures with a transformer-based speech recognition model, which more effectively captures long-range dependencies and leverages full utterance-level context. Second, we introduce a large language model (LLM) as a post-processing module to refine the transcribed text, correcting grammatical inconsistencies and resolving linguistic ambiguities. By integrating contextual modeling at both the acoustic and language levels, the system aims to produce outputs that are intelligible, fluent, and robust enough for real-world deployment.

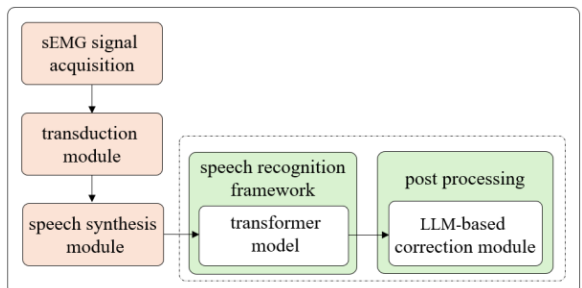

Fig 1. Overview of the proposed SSI recognition framework. The system integrates transformer-based acoustic modeling with an LLM-based post-processing stage for linguistic refinement.

## 2. METHODOLOGY

### 2.1. System Overview

Our work builds upon the silent speech pipeline proposed by Gaddy, which maps surface electromyography (sEMG) signals to intelligible acoustic speech. The baseline framework includes signal preprocessing, feature extraction, EMG-to-acoustic transduction with temporal alignment, and waveform reconstruction using a neural vocoder. We extend this pipeline by integrating a recognition and correction stage to improve intelligibility in downstream tasks.

### 2.2. Dataset

All experiments are conducted on the Digital Voicing dataset introduced by Gaddy. This publicly available corpus contains parallel recordings of vocalized and silent speech from multiple speakers, acquired using an 8-channel surface EMG (sEMG) setup sampled at 1000 Hz. Designed specifically for EMG-to-speech transduction, the dataset provides aligned articulatory signals and corresponding audio targets, enabling objective evaluation of transduction quality through waveform synthesis and downstream ASR performance.

### 2.3. Audio Target Transfer, Transduction, and Speech Synthesis

Each sEMG channel is band-limited with a triangular filter (≈115 Hz cutoff), segmented into 31 ms frames with an 11.6 ms stride, and represented by five time-domain descriptors—root mean square (RMS), mean, energy, absolute value, and zero-crossing rate—together with magnitude coefficients from a 16-point short-time Fourier transform (STFT). Concatenating across the eight channels yields a 112-dimensional feature vector per frame, providing a compact representation of articulatory activity for acoustic mapping.

The EMG-to-speech transduction module maps these features into 80-band mel-spectrograms using a Transformer encoder, following Gaddy's framework. The model consists of six Transformer layers with multi-head self-attention (8 heads, model dimension 768, feedforward dimension 3072, dropout 0.2) and relative positional embeddings spanning ±100 frames. Training is performed with Euclidean distance loss between predicted and reference mel-spectrograms, enabling the model to learn robust correspondences between articulatory and acoustic domains.

Following Gaddy, we adopt the Audio Target Transfer (ATT) strategy to address the absence of acoustic targets in silent speech. In this approach, silent EMG ($Es$) is aligned with vocalized EMG ($Ev$) using dynamic time warping (DTW), refined with canonical correlation analysis (CCA), and paired with the corresponding vocalized audio ($Av$) to provide pseudo-targets for training. This allows the model to learn mappings for both silent ($Es \rightarrow$ pseudo-$Av$) and vocalized ($Ev \rightarrow Av$) conditions.

For waveform reconstruction, HiFi-GAN, a GAN-based vocoder, is utilized to generate high-fidelity waveforms from predicted mel-spectrograms. Compared to autoregressive alternatives such as WaveNet, HiFi-GAN offers significantly faster inference while maintaining perceptual quality, making it well-suited for real-time SSI applications.

### 2.4. Enhanced Speech Transcription with Transformers

The baseline system employs DeepSpeech, an RNN-based ASR framework with LSTM layers for transcribing the generated speech waveform. While RNNs are effective at modeling temporal dependencies through sequential processing, they often degrade in performance when faced with noisy or ambiguous inputs—a frequent challenge in silent speech interfaces, where similar muscle activations can correspond to different phonemes. This limitation motivates the use of models that capture context across the entire utterance rather than relying solely on local sequential dynamics.

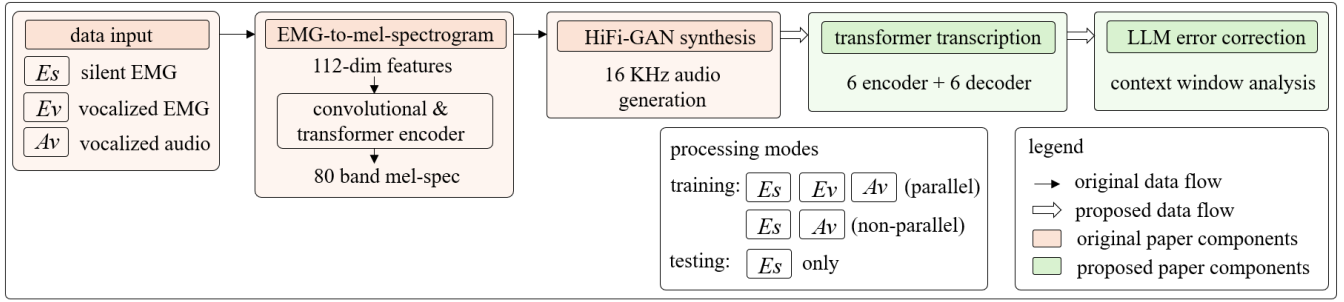

Fig.2 Detailed representation of the proposed dual-stage silent speech recognition framework. The baseline EMG-to-speech pipeline (orange) consists of feature extraction, EMG-to-mel-spectrogram transduction, and waveform synthesis with HiFi-GAN. Our contribution (green) introduces a transformer-based transcription module and an LLM-based error correction stage to improve intelligibility. Training uses both silent and vocalized EMG with audio (parallel and non-parallel), while inference relies on silent EMG alone.

To address this, we adopt a Transformer-based transcription model, building on the self-attention mechanism introduced by Vaswani et al. Unlike RNNs, Transformers process input tokens in parallel, allowing each position in the sequence to attend to all others. This global context modeling is particularly advantageous for SSI, where synthesized speech signals may lack clear phonetic boundaries. By leveraging long-range dependencies across the utterance, the Transformer architecture offers a more robust alternative for transcription in silent speech settings compared to RNN-based baselines.

The Transformer model introduced by Vaswani et al. processes all input tokens in parallel using self-attention, in contrast to RNNs, which operate sequentially. This mechanism allows each position in the input to attend to all others, capturing long-range dependencies across the sequence. Such context modeling is particularly beneficial for ASR in SSI, where synthesized speech signals are often noisy and ambiguous.

Our transcription model follows the standard Transformer architecture with six encoder and six decoder layers, each comprising multi-head self-attention and feedforward sublayers. Within each self-attention block, queries $Q$, keys $K$, and values $V$ are computed from the input features and combined as

$$Attention(Q,K,V)=SoftMax(QK^T/\sqrt{d_k})V \quad (1)$$

where $d_k$ is the key dimensionality, ensuring proper scaling for gradient stability.

The audio waveform is first transformed into log-mel spectrograms via

$$M_t=log(MelFilterBank(|STFT(x_t)|^2)) \quad (2)$$

which provides a compact time-frequency representation retaining essential speech cues while suppressing irrelevant variability. Decoding employs beam search to consider multiple candidate sequences, reducing the likelihood of transient errors compared to greedy decoding.

A key advantage of the Transformer in this setting is its robustness to noisy, imperfectly generated inputs, such as waveforms synthesized from EMG signals that may lack voicing or exhibit misalignment. The attention mechanism allows the model to down-weight corrupted frames while emphasizing clearer regions, thereby improving recognition accuracy. Multi-head attention further enriches this process by enabling different heads to capture complementary dependencies: some attend to short-term local acoustic cues, while others capture long-range relationships. By integrating local and global patterns, the Transformer can disambiguate unclear segments and infer missing phonetic information, ultimately producing more reliable transcription for SSI-generated speech.

### 2.5. LLM-Based Correction

To further improve transcription quality, we incorporate a Large Language Model (LLM) as a post-processing module operating on the ASR output. While the Transformer ASR provides the initial transcription, residual ambiguities and grammatical inconsistencies remain due to the noisy nature of SSI-generated speech. The LLM leverages its contextual reasoning capability to refine these transcriptions, correcting grammatical errors, resolving incomplete words or phrases, and improving overall fluency.

To maintain semantic reliability, candidate corrections are filtered using conservative constraints: generic substitutions are excluded, trivial edits are removed through minimum-length requirements, and only domain-relevant corrections are retained. A confidence threshold of 0.7 is applied to discard low probability outputs. This ensures that refinements are context-aware and linguistically coherent, while avoiding over-correction. By integrating an LLM-based refinement stage, the system effectively reduces recognition errors and enhances the intelligibility and naturalness of transcribed silent speech.

## 3. EXPERIMENTAL RESULTS

All experiments are conducted using the NVIDIA RTX 6000A GPU under the PyTorch framework. For comparability, we follow the same hyperparameter settings as in the baseline Digital Voicing study. In addition, the proposed model components are trained with their respective optimized hyperparameters: transformer encoder-decoder with 8 attention heads (embedding dimension=512, feedforward network dimension=2048, dropout=0.1), EMG signal processing at 16kHz sampling rate, and decoder beam search width of 500. For the LLM correction component, we use GPT-2 with a max sequence length of 128 tokens and filtering thresholds to ensure domain specificity. Model training employs mean squared error (MSE) loss with the Adam optimizer. To improve robustness across recording conditions, we incorporate 32-dimensional session embeddings, which capture inter-session variability while maintaining consistent transduction performance. Mixed training is performed with both silent and vocalized data, combining parallel and non-parallel alignments, thereby enhancing generalization across speaking modes.

The evaluation was conducted on the complete test set (≈100 utterances) and Table 1 summarizes the results in terms of word error rate (WER), relative improvement over the baseline, and average evaluation time per utterance. The average per-utterance latency varies with utterance length.

Table 1. Comparison of WER, relative improvement, and evaluation time for the baseline RNN-based ASR and the proposed Transformer-based ASR with and without LLM correction.

| System | WER (%) | Relative improvement (%) | Average time taken per utterance (sec) |
|---|---|---|---|
| Deep Speech (RNN-Based ASR baseline) | 36 | Baseline | 1.42 |
| Transformer based ASR (proposed) | 32.5 | 9.7 | 0.73 |
| Transformer + LLM Correction (proposed) | 30 | 16.6 | 0.78 |

Replacing the RNN-based DeepSpeech recognizer with a Transformer model reduced the WER to 32.5%, corresponding to a 9.7% relative improvement over the 36% baseline. Incorporating the LLM-based correction module further reduced WER to 30%, yielding a 16.6% relative improvement compared to baseline. Taken together, this dual enhancement process achieves a 6% absolute gain, marking a significant step toward unambiguous, clear, and fluent transcription in silent speech interfaces.

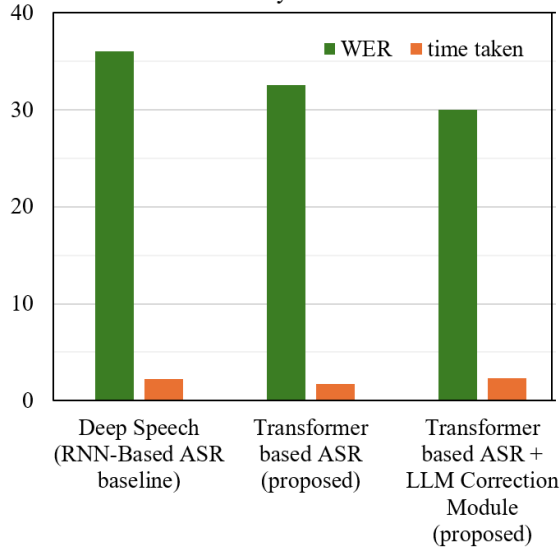

Fig. 3 Visualization of WER and evaluation time across systems: baseline RNN-based ASR (DeepSpeech), Transformer-based ASR, and Transformer with LLM correction.

The improvements, as visualized in Fig. 3, can be attributed to two complementary factors. First, the Transformer-based ASR overcomes the limitations of RNNs by using self-attention to model utterance-level dependencies in parallel, making it more robust to the phonetically ambiguous and noisy waveforms synthesized from EMG signals. This yields a clear WER reduction compared to the RNN baseline. Second, the LLM correction module provides additional refinement at the linguistic level. Even after acoustic modeling, errors remain due to overlapping EMG activations and imperfect waveform synthesis. By applying context-aware post-processing, the LLM corrects grammatical inconsistencies, restores incomplete words, and enforces semantic fluency. A conservative filtering mechanism ensures that only high-confidence corrections are applied, avoiding overcorrection while producing more coherent transcripts. This explains the further 2.5% absolute reduction in WER beyond the Transformer-only model.

In terms of runtime, the Transformer is also more efficient than the RNN baseline, reducing evaluation time from 1.42 to 0.73 seconds as the average per utterance. This efficiency stems from the parallelizable nature of self-attention compared to recurrent updates. Adding the LLM introduces modest overhead, raising the total to 0.78 seconds, which remains practical given the intelligibility gains. On a per-sentence basis, inference requires only a few seconds depending on utterance length, making the system feasible for near real-time applications.

## 4. LIMITATIONS, FUTURE DIRECTIONS, CONCLUSION

This paper proposed a dual-stage pipeline for silent speech recognition, combining a Transformer-based ASR for accurate transcription with an LLM-based module for linguistic refinement. Quantitative evaluation demonstrated a consistent reduction in WER across both stages, confirming the effectiveness of integrating context-aware acoustic modeling with language-level correction. Together, these improvements achieved a 6% absolute gain over the baseline, advancing the intelligibility and fluency of SSI outputs.

Despite these gains, certain limitations remain. The synthesized speech, while intelligible, can still exhibit disfluencies and occasional unnatural phrasing. Our current system mitigates this with conservative correction strategies, but more robust solutions will be needed as we target deployment in broader scenarios. Another limitation is that the LLM has been primarily constrained to error correction; its full potential for free-form correction and adaptation remains underexplored.

Future work will extend the role of LLMs beyond grammatical correction to encompass style adaptation, personalization, and dialectal variation. We also plan to explore optimization strategies for lightweight and resilient deployment, enabling real-time SSI decoding on resource-constrained devices. These directions will support broader applicability in assistive communication and everyday human–machine interaction.